\documentclass[runningheads]{llncs}
\usepackage{graphicx,xcolor}
\usepackage{comment}
\usepackage{caption}
\usepackage[export]{adjustbox}
\usepackage{here}
\begin{document}
\title{Character-independent font identification}
\author{Daichi Haraguchi
\and
Shota Harada
\and
Brian Kenji Iwana\orcidID{0000-0002-5146-6818}
\and
Yuto Shinahara 
\and
Seiichi Uchida\orcidID{0000-0001-8592-7566}
}
\authorrunning{Haraguchi et al.}
\institute{Kyushu University, Fukuoka, Japan\\
\email{\{daichi.haraguchi, shota.harada, brian, uchida\}@human.ait.kyushu-u.ac.jp}
}
\maketitle 
\begin{abstract}
There are a countless number of fonts with various shapes and styles. In addition, there are many fonts that only have subtle differences in features. Due to this, font identification is a difficult task. In this paper, we propose a method of determining if any two characters are from the same font or not. This is difficult due to the difference between fonts typically being smaller than the difference between alphabet classes. Additionally, the proposed method can be used with fonts regardless of whether they exist in the training or not. In order to accomplish this, we use a Convolutional Neural Network (CNN) trained with various font image pairs. In the experiment, the network is trained on image pairs of various fonts. We then evaluate the model on a different set of fonts that are unseen by the network. The evaluation is performed with an accuracy of 92.27\%. Moreover, we analyzed the relationship between character classes and font identification accuracy. 

\keywords{Font identification \and Representation learning \and  Convolutional neural networks.}
\end{abstract}

\section{Introduction\label{sec:intro}}

\begin{figure}
    \begin{center}
 \begin{minipage}{0.5\hsize}
  \begin{center}
   (a)\quad\includegraphics[width=0.8\textwidth,valign=c]{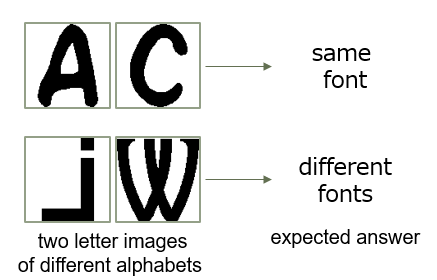}
  \end{center}
  \label{fig:one}
\end{minipage}\\
(b)\quad\includegraphics[width=0.8\textwidth,valign=c]{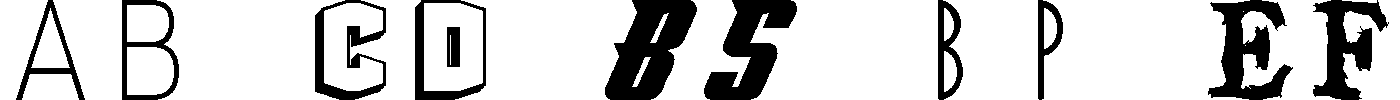}\\[2mm]
(c)\quad\includegraphics[width=0.8\textwidth,valign=c]{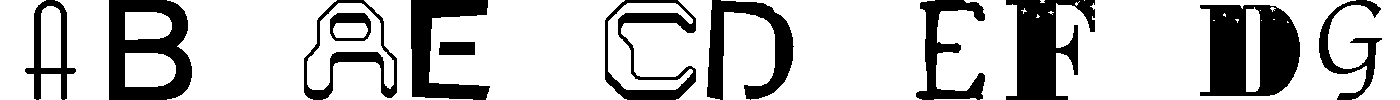}\\[2mm]
(d)\quad\includegraphics[width=0.8\textwidth,valign=c]{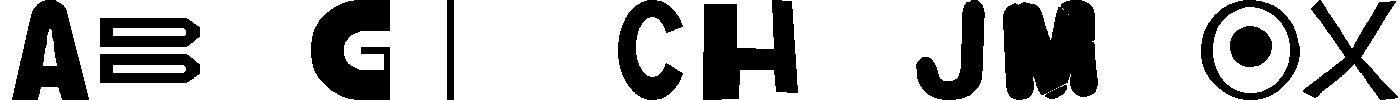}\\[2mm]
(e)\quad\includegraphics[width=0.8\textwidth,valign=c]{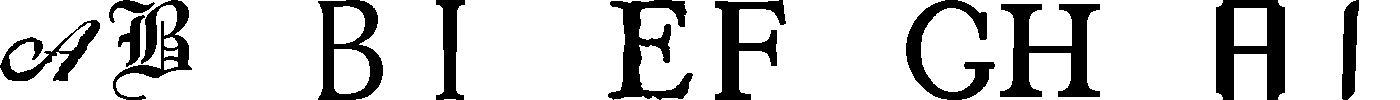}\\
        \caption{Explanation of our task and examples of character image pairs from different classes. (a) is an explanation of our task. The pairs in, (b) shows the same font pairs, whereas (c) shows different font pairs.  In contrast, (d) and (e) are difficult cases; (d) shows the same pairs, whereas (e) shows different pairs.}
        \label{fig:task-examples}

    \end{center}
\end{figure}

In this paper, we tackle font identification from different character classes. Specifically, given a pair of character images from different character classes\footnote{Throughout this paper, we assume that the pairs come from different character classes. This is simply because our font identification becomes a trivial task for the pairs of the same character class  (e.g., `A'); if the two images are exactly the same, they are the same font; otherwise, they are different.
}, we try to discriminate
whether the images come from the same font or not. Fig.~\ref{fig:task-examples}~(a) shows example input pairs of the task; we need to the same or different font pairs. It is easy to identify that Fig.~\ref{fig:task-examples}~(b) is the same font pair. It is also easy to identify that Fig.~\ref{fig:task-examples}~(c) is a different font pair. In contrast, the examples in Fig.~\ref{fig:task-examples}~(d) and (e) are more difficult. Fig.~\ref{fig:task-examples}~(d) shows the same font pairs, whereas (e) shows different font pairs.
\par

This task is very different from the traditional font identification task, such as \cite{wang2015deepfont,ma2005font,gupta2016font}. In the traditional task, given a character image (a single letter image or a single word image or a sentence image), we need to answer its font name (e.g., Helvetica). In a sense, it is rather a recognition task than an identification task. This is because, in the traditional task, we only can identify the fonts which are registered in the system in advance. In other words, it is a multi-class font recognition task and each class corresponds to a known font name. 
In contrast, our font identification task is a two-class task to decide whether a pair of character images come from the same font or different fonts, 
without knowing those font names beforehand. 
 \par 
%
We propose a system for our font identification task for a pair of character images from different character classes. The system is practically useful because the system will have more flexibility than the systems for the traditional font identification task. As noted above, the traditional systems only ``recognize'' the input image as one of the fonts that are known to the system. However, it is impossible to register all fonts to the system because new fonts are generated everyday in the world. (In the future, the variations of fonts will become almost infinite since many automatic font generation systems have been proposed, such as ~\cite{gan,suveeranont2010example,li2018bezier,miyazaki2019automatic}.) Accordingly, the traditional systems will have a limitation on dealing with those fonts that are ``unknown'' to them. Since the proposed system does not assume any font class, it can deal with arbitrary font images. \par
Moreover, since the proposed system assumes single character images as its 
input, we can perform font identification even if a document contains a small number of characters. For example, analysis of incunabula or other printed historical documents often needs to identify whether two pieces of documents are printed in the same font or not. A similar font identification task from a limited number of characters can be found in forensic researches. For example, forensic experts need to determine whether two pieces of documents are printed by the same printer or not.
\par
%

\begin{figure}[t]
    \begin{center}
		\includegraphics[width=0.8\textwidth]{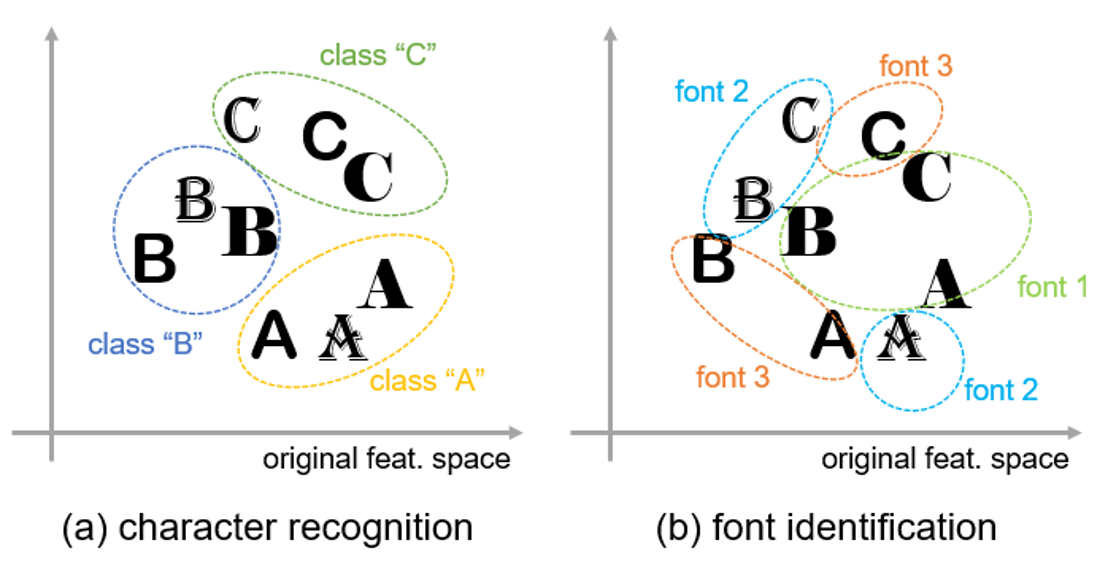}
		\caption{Comparing multi-font character recognition~(a), our font identification~(b) is a more difficult classification task.}
		\label{fig:difficulty}
	\end{center}
\end{figure}

In addition to the above practical merits, our font identification is a challenging scientific task. In fact, our task is very difficult even though it is formulated just as a binary classification problem. Fig.~\ref{fig:difficulty} illustrates the distribution of image samples in a feature space. As a success of multi-font optical character recognition~(OCR)~\cite{Uchida} proves, the samples from the same character class form a cluster and the clusters of different character classes are distant in the feature space.
This is because inter-class variance is much larger than intra-class variance; that is, the difference by the character classes is larger than the difference by the fonts. This fact can be confirmed by imagining the template matching-based identification. Although we can judge the class identity of two images (in different fonts) even by template matching, we totally cannot judge the font identity of two images (in different character classes). Consequently, our system needs to disregard large differences between character classes and emphasize tiny differences (such as the presence or absence of serif) in fonts. We find a similar requirement in the text-independent writer identification task, such as \cite{nguyen2019text}.
\par

In this paper, we experimentally show that even a simple two-stream convolutional neural network (CNN) can achieve a high accuracy for our font identification task, in spite of the above anticipated difficulty. Our CNN is not very modern (like a CNN with a feature disentanglement function~\cite{liu2018exploring,liu2018unified,press2018emerging}) but simply accepts two-character image inputs and makes a decision for the binary classification (i.e., the same font or not). In addition, we show - detailed analysis of the identification results. For example, we will observe which alphabet pairs (e.g., `A'-`K') are easier or more difficult for the identification. In fact, there is a large difference in the identification performance among alphabet pairs.  
\par

The main contributions of our paper are summarized as follows:
\begin{itemize}
    \item To the authors' best knowledge, this is the first attempt of font identification for different character classes. 
    \item Through a large-scale experiment with more than 6,000 different fonts, we prove that even a simple two-stream CNN can judge whether two-letter images come from the same font or not with a high accuracy ($>90\%$), in spite of the essential difficulty of the task. It is also experimentally shown that the trained CNN has a generalization ability. This means that the representation learning by the simple CNN is enough to extract font style features while disregarding the shape of the character class.  
    \item By analyzing the experimental results, we proved the identification accuracy depends on character class pairs. For example, `R' and `U' are a class pair with a high accuracy, whereas `I' and `Z' are with a lower accuracy.
\end{itemize}

\section{Related Work\label{sec:related}}
\subsection{Font Identification and Recognition}

To the authors' best knowledge, this is the first trial of font identification in our difficult task setting. 
Most of past research on font identification is visual font recognition where a set of fonts are registered with their names and an input character image is classified into one of those font classes. 
These systems traditionally use visual features extracted from characters. 
For example, Ma and Doermann~\cite{ma2005font} use a grating cell operator for feature extraction and Chen et al.~\cite{chen} use a local feature embedding. 
In addition, visual font recognition has been used for text across different mediums, such as historical documents~\cite{gupta2016font} and natural scene text~\cite{chen}. 
Font recognition has also been used for non-Latin characters, such as Hindi~\cite{bagoriya_2014}, 
Farsi~\cite{Khosravi_2010}, 
Arabic~\cite{Ben_Moussa_2010,M__2017}, 
Korean~\cite{jeong2003identification}, 
Chinese~\cite{Yang_2006},
etc. 
Recently, neural networks have been used for font identification. 
DeepFont~\cite{wang2015deepfont} uses a CNN-based architecture for font classification.  

However, these font identification methods classify fonts based on a set number of known fonts. 
In contrast, the proposed method detects whether the fonts are from the same class or not, independent from what fonts it has seen. 
This means that the proposed method can be used for fonts that are not in the dataset, which can be an issue given the growing popularity of font generation~\cite{abe2017font,gan,miyazaki2019automatic}. 

In order to overcome this, an alternative approach would be to only detect particular typographical features or groups of fonts. 
Many classical font recognition models use this approach and detect typographical features such as typeface, weight, slope, and size~\cite{Min_Chul_Jung_1999,Chaudhuri,shinahara2019serif}. 
In addition, clustering has been used to recognize groups of fonts
\cite{O_ztu_rk_2001,Avil_s_Cruz_2004}.

\subsection{Other Related Identification Systems}

The task of font identification can be considered as a subset of script identification. 
Script identification is a well-established field that-aims to recognize the script of text, namely, the set of characters used. 
In general, these methods are designed to recognize the language for individual writing-system OCR modules~\cite{ghosh2010script}. 
Similar to font identification, traditional script identification use visual features such as Gabor filters~\cite{Tan_1998,Pan_2005} and text features~\cite{Elgammal,Pal_2002}. 

Furthermore, font identification is related to the field of signature verification and writer identification. 
In particular, the task of the proposed method is similar to writer-independent signature verification in that both determine if the text is of the same source or different sources. 
Notably, there are methods in recent times which use CNNs~\cite{Hafemann_2017,zheng2019capturing} and Siamese networks~\cite{Xing_2018,Ruiz_2020} that resemble the proposed method. 

\section{Font Identification by Convolutional Neural Networks\label{sec:method}}

Given a pair of character images $\mathbf{x}_c$ and $\mathbf{x}_d$ of font class $c$ and $d$ respectively, our task is to determine if the pair of characters are of the same font ($c=d$) or different fonts ($c\neq d$). 
In this way, the classifier assigns a binary label indicating a positive match and a negative match. 
The binary label is irrespective of the character or actual font of the character used as an input pair. 

\begin{figure}[t]
	\begin{center}
	\includegraphics[scale=0.30]{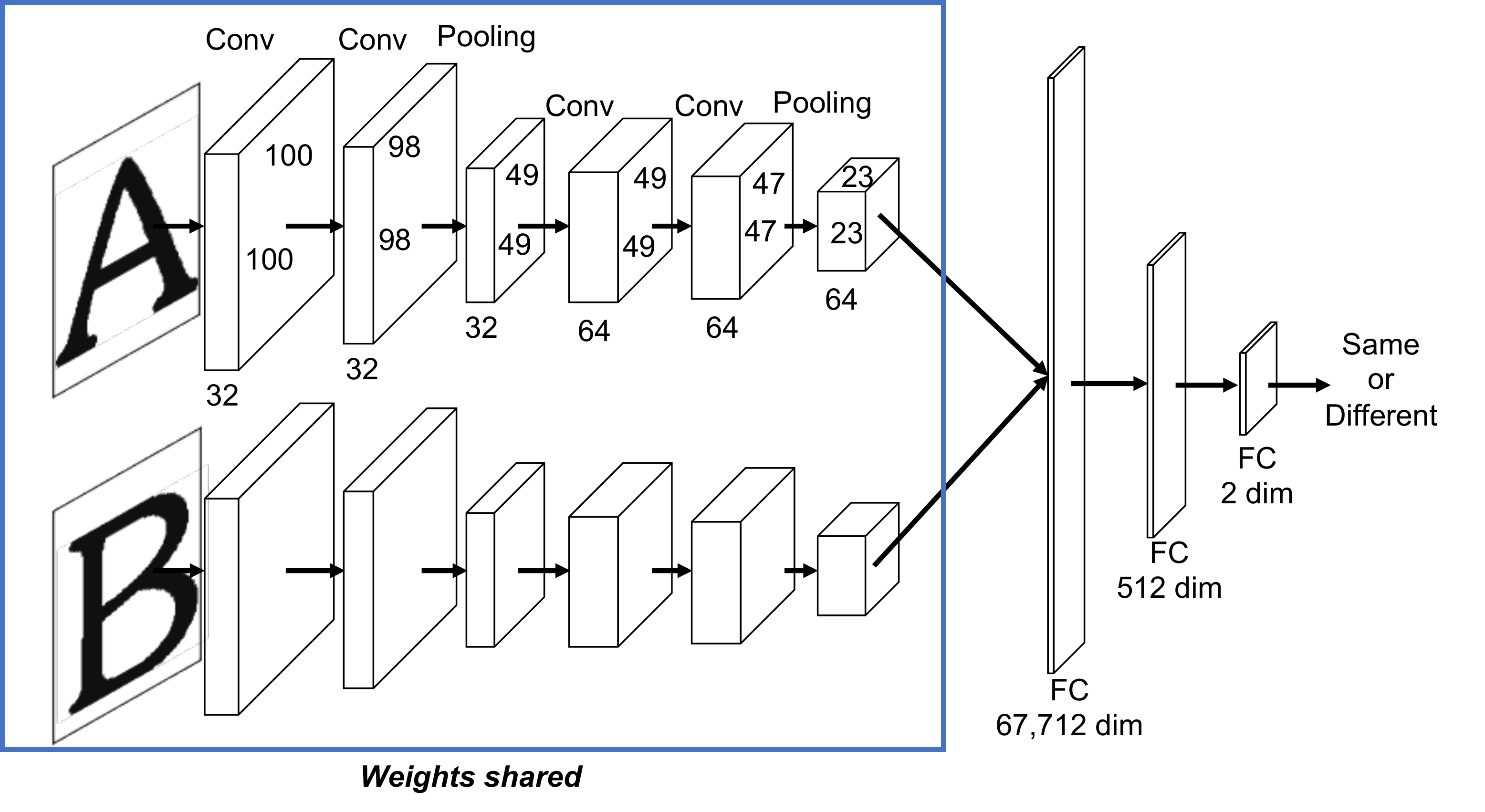}
		\caption{Structure of the neural networks for font identification.}
		\label{fig:network-structure}
	\end{center}
\end{figure}

In order to perform the font identification, we propose a two-stream CNN-based model. 
As shown in Fig.~\ref{fig:network-structure}, a pair of input characters are fed to separate streams of convolutional layers which are followed by fully-connected layers and then the binary classifier. 
In addition, the two streams of convolutional layers have the same structure and shared weights. 
This is similar to a Siamese network~\cite{koch2015siamese}, typically used for metric learning, due to the shared weights. 
However, it differs in that we use a binary classifier with cross-entropy loss. 

Each stream is comprised of four convolutional layers and two max pooling layers. 
The kernel size of the convolutions is $3 \times 3$ with stride 1 and the kernel size of the pooling layers is $2 \times 2$ with stride 2.
The features from the convolutional layers are concatenated and fed into three fully-connected layers. 
Rectified Linear Unit (ReLU) activations are used for the hidden layers and softmax is used for the output layer. 
During training, dropout with a keep probability of 0.5 is used after the first pooling layer and between the fully-connected layers. 


\section{Experimental Results\label{sec:experiment}}


\begin{table}[t]
\caption{Confusion matrix of the test set\label{table:accuracy}}
    \centering
    \begin{tabular}{c|r|r}
      GT\textbackslash predicted   & 
      \multicolumn{1}{c}{same} & \multicolumn{1}{|c}{different}\\ \hline
      same   & $196,868 \pm1,758$  & $7,232 \pm1,758$ \\  \hline
        different  & $24,331 \pm2,014$ & $179,769 \pm2,014$\\ 
    \end{tabular}
\end{table}


\begin{figure}[t]
\begin{center}
    \includegraphics[scale=0.4]{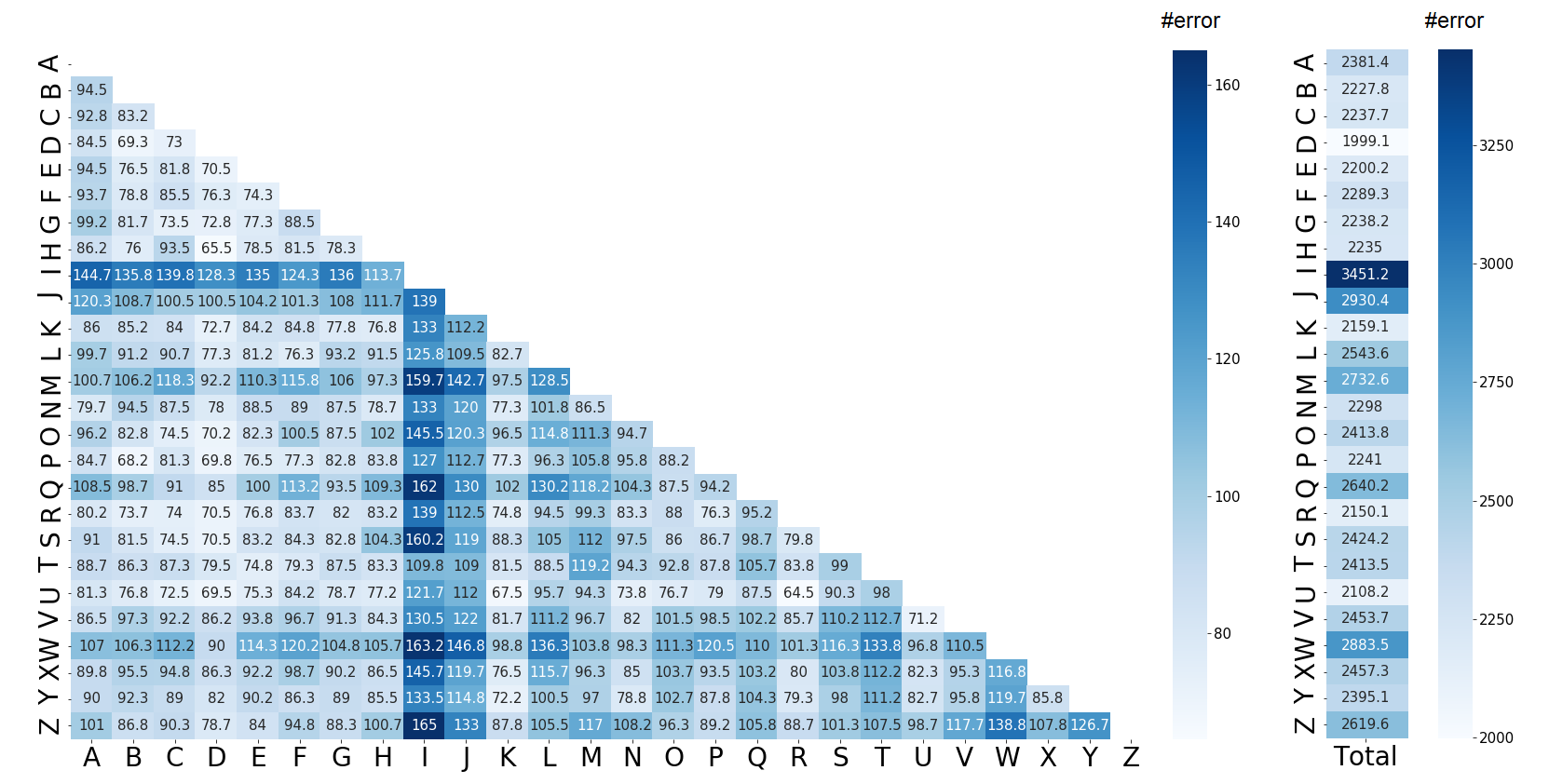}\\
    \caption{Misidentification by class}
	\label{fig:misidentification_matrix}
\end{center}
\end{figure}

\subsection{Font Dataset\label{sec:dataset}}

The dataset used for the experiment were 6,628 fonts from the Ultimate Font Download\footnote{\url{http://www.ultimatefontdownload.com/}}. 
Although the total font package is originally comprised of about 11,000, we removed ``dingbat'' fonts (i.e., icon-like illustrations and illegible fonts) for the experiments and the 6,628 fonts remain. This font dataset still contains main fancy fonts; in appendix, we will discuss another dataset with more formal fonts. 
To construct the dataset, we rasterize the 26 uppercase alphabet characters into $100 \times 100$ binary images. 
We only use uppercase characters in this paper for experimental simplicity. 
Although, it should be noted that there are some fonts that contain lowercase character shapes as uppercase characters. 



The 6,628 fonts were divided into three font-independent sets, 5,000 for training, 1,000 for validation, and 628 for training. 
Within each set, we generated uppercase alphabet pairs from the same font (positive pairs) and different fonts (negative pairs). 
Each of the pairs uses different alphabetical characters. 
Furthermore, each combination of characters is only used one time, i.e. either A'-`B' or `B'-`A' is used, but not both. 
Therefore, we have $_{26}C_2 = 325$ total pairs of each font. 
Consequently, the training set has $5,000\times 325 \approx 1.60\times 10^6$ positive pairs. 
An equal number of negative pairs are generated by randomly selecting fonts within the training set. 
Using this scheme, we also generated $2 \times 3.25\times 10^5$ for validation and approximately $2 \times 2.04\times 10^5$ for testing. 
In addition, as outlined in Appendix~\ref{app:adobe}, a second experiment was performed on an external dataset to show the generalization ability of the trained model on other fonts.

\begin{figure}[t]
\begin{center}
    \includegraphics[scale=0.4]{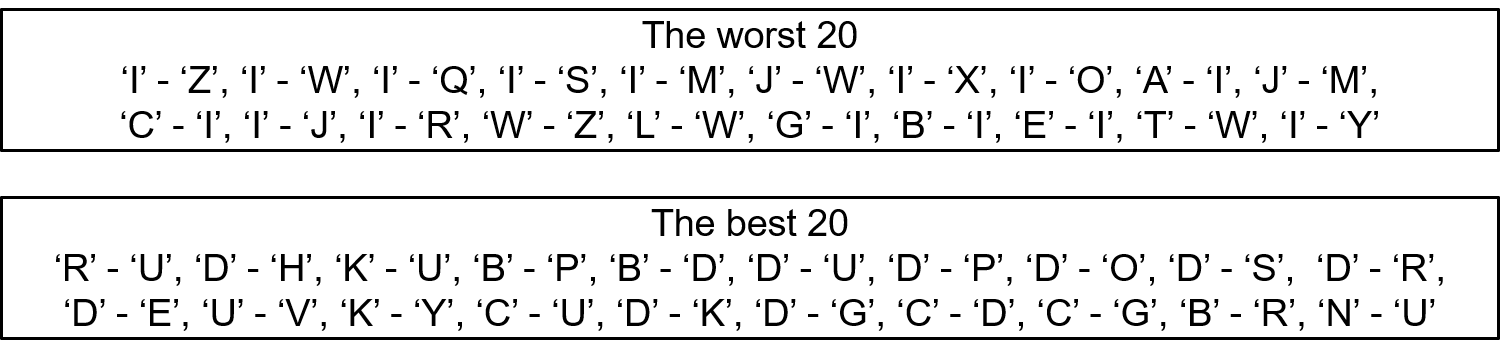}\\
    \caption{The character pairs with the 20 worst and 20 best accuracies}
	\label{fig:misidentification_list}
\end{center}
\end{figure}

\subsection{Quantitative Evaluation\label{sec:quantitative}}

We conduct 6-fold cross-validation to evaluate the accuracy of the proposed CNN. 
The identification accuracy for the test set was $92.27 \pm{0.20}$\%. 
The high accuracy demonstrates that it is possible for the proposed method to determine if the characters come from the same font or not, even when they come from different characters. 
Table~\ref{table:accuracy} shows a confusion matrix of the test results. 
From this table, it can be seen that different font pairs have more errors than the same font pairs. 
This means that similar but different font pairs are often misidentified as the same font.


Among the misidentification, there are some character pairs that are more difficult to classify than others. 
As shown in Fig.~\ref{fig:misidentification_matrix}, we find that the pairs including `I' or `J' are more difficult. 
This is due to there being very little differences in visible features in `I' and `J' due to their simplicity. 

Additionally, we found that character pairs with similar features are predictably easier to differentiate and character pairs with different features are difficult. 
In other words, the amount of information that characters have, such as angles or curves, is important for separating matching fonts and different fonts. 
For example, in Fig.~\ref{fig:misidentification_matrix}, the number of misidentification of the `I'-`T' pair is the lowest of any pair including and `I' because `T' has the most similar shape to 'I'. 
We also find that the number of misidentifications for `D', `K', `R', and `U' are the least because they have the most representative features of straight lines, curves, or angles. 

This is consistent with other characters with similar features. 
The character pair with the worst classification rate is `I'-`Z' and the character pair with the highest accuracy is `R'-`U,' as outlined in Fig.~\ref{fig:misidentification_list}. 
From this figure, we can see that many characters with similar features have high accuracies. 
For example, `B'-`P,' `B'-`D,' and `O'-`D.' 
As a whole, `C'-`G' and `U'-`V' pairs have fonts that are easy to identify. 
These pairs are not likely to be affected by the shape of the characters. 

Interestingly, the top 5 easiest characters paired with `B' for font identification are `P,' `D,' `R,' `H,' and `E' and the top 5 for `P' are `B,' `D,' `R,' `E,' and `F.' 
In contrast, the top 5 easiest font identifications with `R' are `U,' `D,' `B,' `C,' and `K.' 
`B' and `P' have the same tendency when identifying fonts. 
However, font identification with `R' seems to use different characteristics despite `B,' `P,' and `R' having similar shapes. 
This is because `B' and `P' are composed of the same elements, curves and a vertical line, whereas `R' has an additional component.

\begin{figure}[t]
	\begin{center}
	\includegraphics[width=0.8\textwidth]{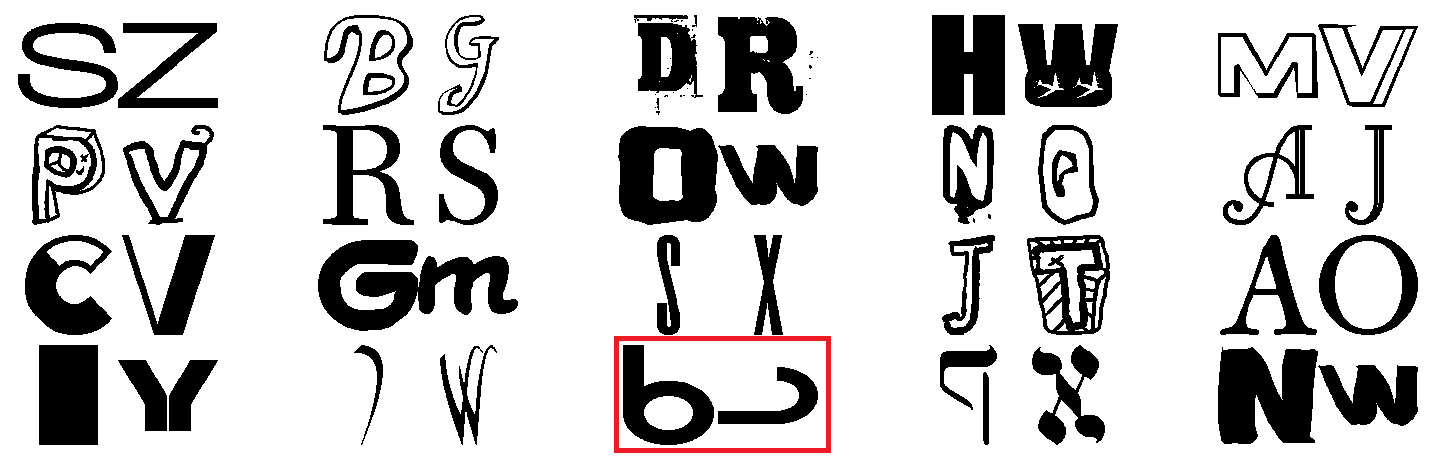}\\[-2mm]
		\caption{Examples of correctly identified pairs (GT: same $\to$ prediction: same). The font pair marked by the red box has a nonstandard character.}
		\label{fig:correct}
	\end{center}
\end{figure}

\begin{figure}[t]
	\begin{center}
	\includegraphics[width=0.8\textwidth]{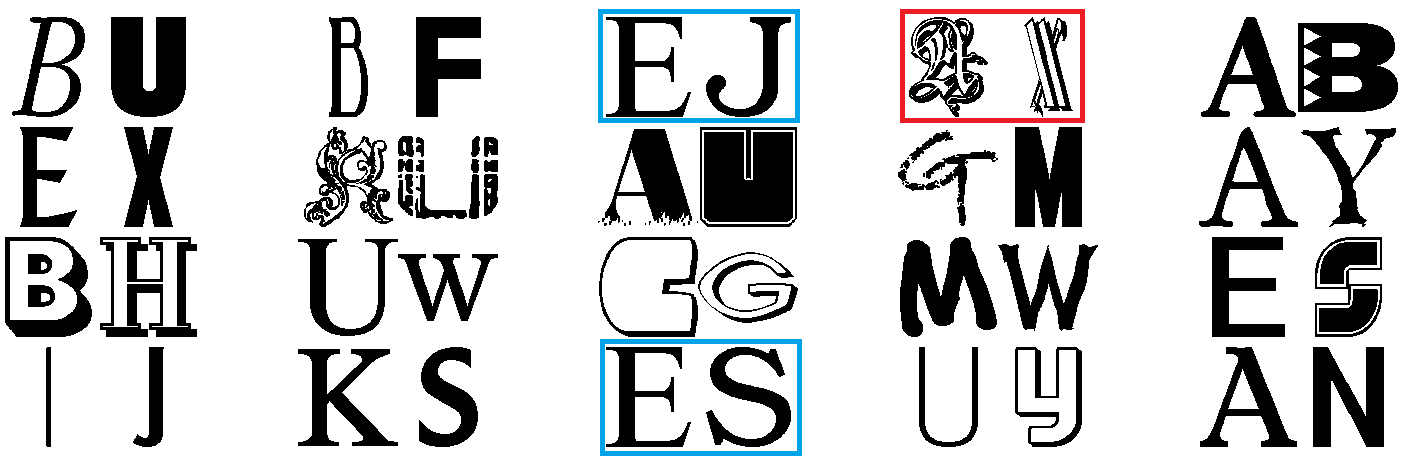}\\[-2mm]
		\caption{Examples of correctly identified pairs (GT: different $\to$ prediction: different). 
		The font pairs marked by the blue boxes have similar but different fonts. The red box indicates fonts that are almost illegible.}
		\label{fig:correct_diff}
	\end{center}
\end{figure}

\subsection{Qualitative Evaluation\label{sec:qualitative}}

We show some examples of correctly identified pairs in Fig.~\ref{fig:correct}.
In the figure, the proposed method is able to identify fonts despite having dramatically different features such as different character sizes. 
However, the font weight of the correctly identified fonts tends to be similar. 
Also notably, in Fig.~\ref{fig:correct}, in the `A'-`O' pair, the `O' does not have a serif, yet, the proposed method is able to identify them as a match. 
Furthermore, the character pair highlighted by a red box in Fig.~\ref{fig:correct} is identified correctly. 
This is surprising due to the second character is unidentifiable and not typical of any character. 
This reinforces that the matching fonts are determined heavily by font weight.

It is also easy for the proposed method to correctly identify different font pairs that have obviously different features to each other.
Examples of different font pairs that are correctly identified are shown in Fig.~\ref{fig:correct_diff}. 
Almost all of the pairs have different features like different line weights or the presence of serif. 
On the other hand, the proposed method was also able to distinguish fonts that are similar, such as `E'-`J' and `E'-`S,' highlighted by blue boxes. 

\begin{figure}[t]
	\begin{center}
	\includegraphics[width=0.8\textwidth]{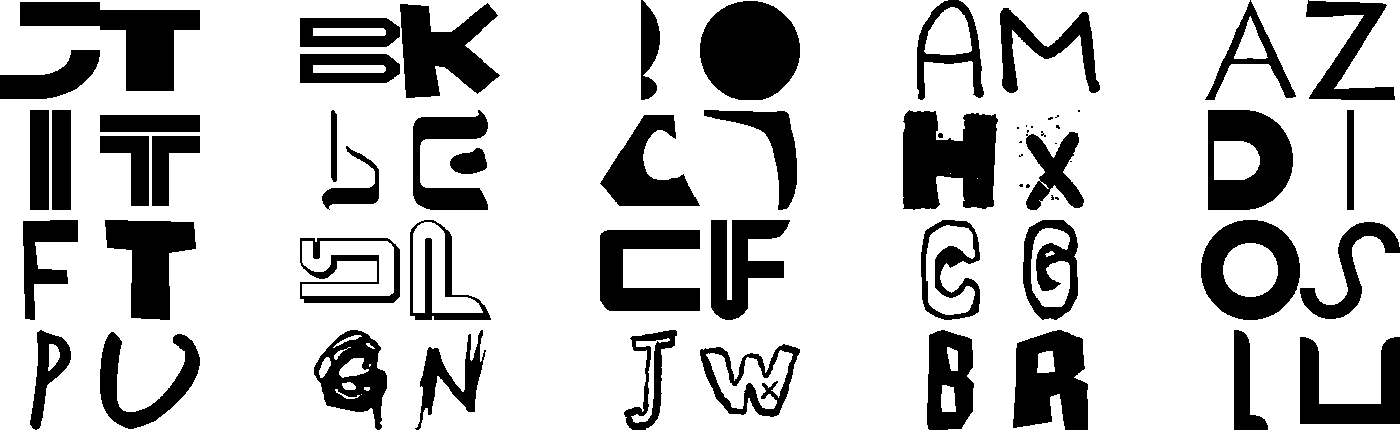}\\[-2mm]
		\caption{Examples of misidentified pairs (GT: same $\to$ prediction: different).}
		\label{fig:confusing-a}
	\end{center}
\end{figure}

\begin{figure}[t]
	\begin{center}
		\includegraphics[width=0.8\textwidth]{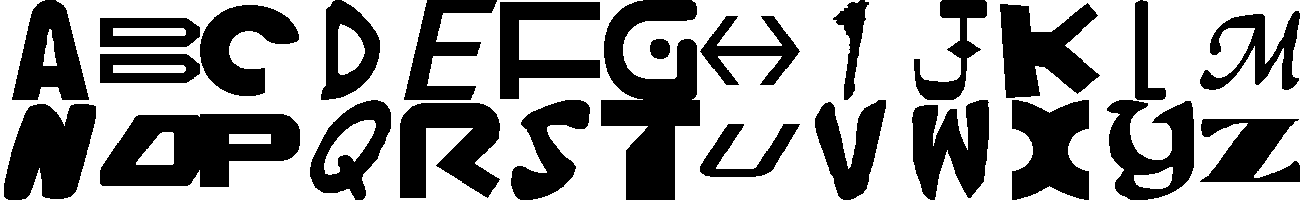}\\[-2mm]
		\caption{The font with the most identification errors (GT: same $\to$ prediction: different).}
		\label{fig:confusing-a-font}	
	\end{center}
\end{figure}

There are also many examples of fonts that are difficult with drastic intra-font differences. 
For example, Fig.~\ref{fig:confusing-a} shows examples of fonts that had the same class but predicted to be from different classes. 
Some of these pairs are obviously the same fonts, but most of the pairs have major differences between each other including different line weight and different shapes. 
In particular, the font in Fig.~\ref{fig:confusing-a-font} is difficult as there is seemingly no relation between the characters. 
This font had the lowest accuracy for the proposed method. 

\begin{figure}[t]
	\begin{center}
	\includegraphics[width=0.8\textwidth]{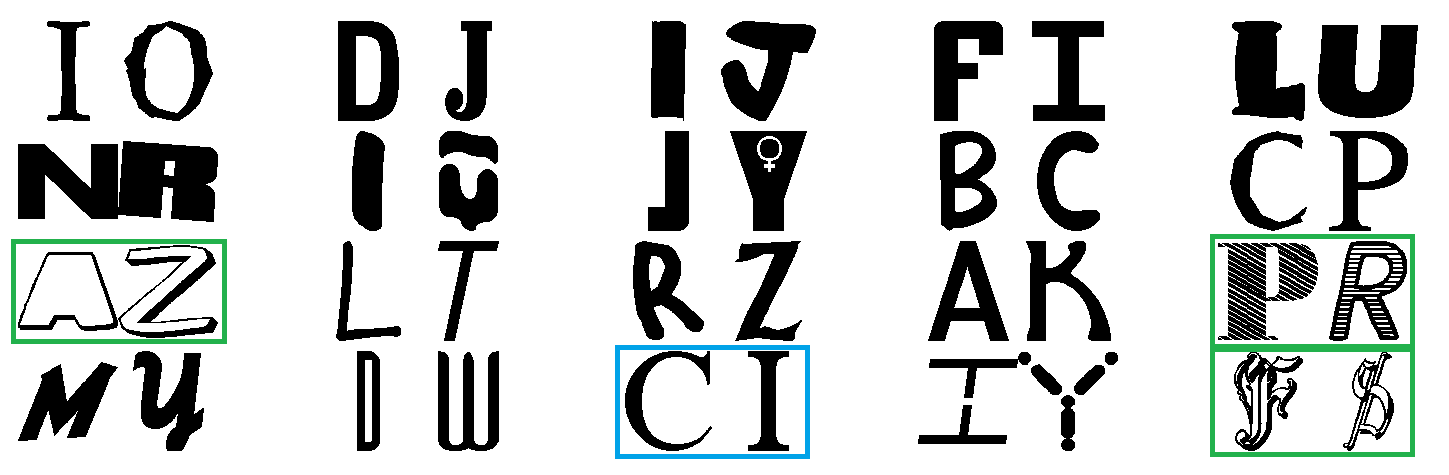}\\[-2mm]
		\caption{Examples of misidentified pairs (GT: different$\to$ prediction: same). The green boxes indicate font pairs which are outlines and the font pair with the blue box is the font is difficult even for humans. }
		\label{fig:confusing-b}

	\end{center}
\end{figure}

There are many fonts that look similar visually but are different which makes it difficult to identify with the proposed method.  
Fig.~\ref{fig:confusing-b} shows examples of font pairs that are misidentified as the same font when they are actually different fonts. 
These fonts are very similar to each other. 
It is also difficult even for us to identify as different. 
Their font pairs have similar features, including line weights, slant lines, and white areas. 

\begin{figure}[t]
    \begin{minipage}{0.5\hsize}
        \begin{center}
            \includegraphics[scale=0.40]{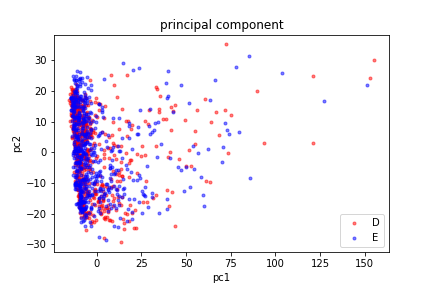}
            \hspace{1.6cm} (a) `D' and `E'.
            \end{center}
    \end{minipage}
    \begin{minipage}{0.5\hsize}
        \begin{center}
            \includegraphics[scale=0.40]{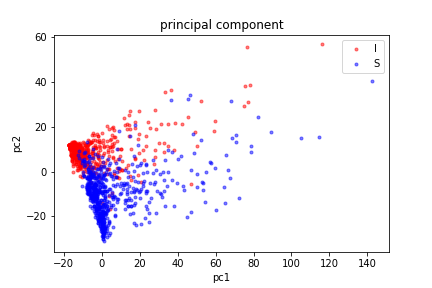}
            \hspace{1.6cm} (b) `I' and `S'.
        \end{center}
    \end{minipage}
	\caption{Visualizing the feature distribution by PCA.} 
	\label{fig:pca}
\end{figure}

\subsection{Font Identification Difficulty by Principal Component Analysis}\label{sec:pca}

We analyze the difference in identification difficulty of font pairs using Principal Component Analysis~(PCA). 
In order to do this, PCA is applied to flattened vectors of the output of each stream. 
Fig.~\ref{fig:pca} shows two character comparisons, `D'-`E' and `I'-`S,' with the test set fonts mapped in a 2D space. 
In the figure, the fonts of the first character are mapped in red and the second character blue, which allows us to compare the similarity of the output of each stream. 
From this figure, we can observe that feature distribution between characters that the proposed method had an easy time identifying, e.g. `D'-`E,' have significant overlap. 
Conversely, characters that were difficult, e.g. `I'-`S,' have very few features that overlap. 
From these figures, we can expect that font identification between characters that contain very different features is difficult for the proposed method.



\subsection{Explanation Using Grad-CAM}\label{sec:gradcam}

We visualize the contribution map of the font pairs toward font identification using Gradient-weighted Class Activation Mapping (Grad-CAM)~\cite{Selvaraju_2019}. 
Grad-CAM is a neural network visualization method which uses the gradient to weight convolutional layers in order to provide instance-wise explanations. 
In this case, we use Grad-CAM to visualize the contribution that regions on the pair of inputs have on the decision process. 
Specifically, the last convolutional layer of each stream is used to visualize the important features of the input. 

\begin{figure}[t]
	\begin{center}
    \begin{minipage}[t]{\hsize}
        \begin{center}
	    \includegraphics[width=1\textwidth]{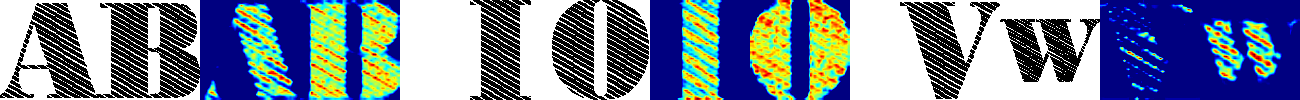}\\
	    {(a) Examples of correctly identified pairs (GT: same $\to$ prediction: same)}\\
	    \end{center}
	\end{minipage}
	\vskip 0.3cm
	\begin{minipage}[t]{\hsize}
	    \begin{center}
	    \includegraphics[width=1\textwidth]{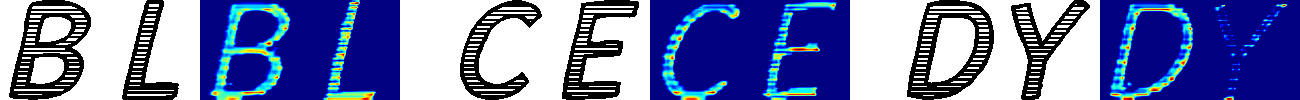}\\
	    {(b) Examples of correctly identified pairs (GT: same $\to$ prediction: same)}\\
	    \end{center}
	\end{minipage}
	\vskip 0.3cm
	\includegraphics[width=1\textwidth]{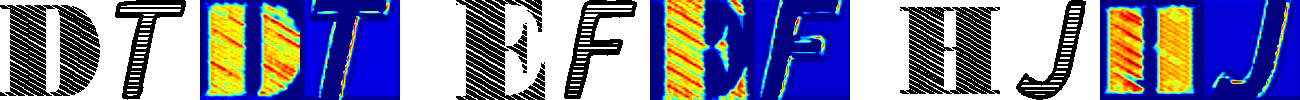}
		{(c) Examples of correctly identified pairs (GT: different $\to$ prediction: different)}\\
	\vskip 0.3cm
	\includegraphics[width=1\textwidth]{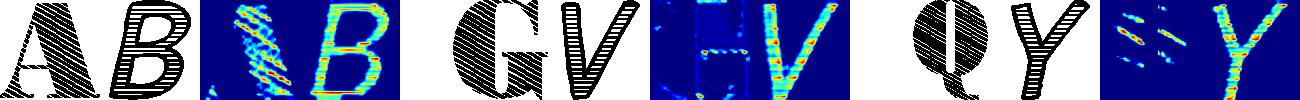}
	    {(d) Examples of incorrectly identified pairs (GT: different $\to$ prediction: same)}\\
	\end{center}
 	\vspace{-0.5cm} 
	\caption{Visualization of difficult pairs using Grad-CAM}
	\label{fig:gradcam_1}
\end{figure}

We first visualize font pairs which are difficult due to having a similar texture fill.
Fig.~\ref{fig:gradcam_1} shows two fonts that were easily confused by the proposed method. 
Fig.~\ref{fig:gradcam_1}~(a) has examples where the first font which were correctly identified as the same and (b) is examples of the second font. 
From these, we can confirm that the presence of the lined fill contributed heavily to the classification. 
Note that even characters with dramatically different shapes like `I'-`O' put a large emphasis on contribution to the filling.


Next, we compared the results of Grad-CAM to correctly identified different pairs. 
The results are shown in Fig.~\ref{fig:gradcam_1}~(c). 
In this case, Grad-CAM revealed that the network focused on the outer regions of the second font. 
This is due to that font containing a subtle outline. 
Accordingly, the network focused more on the interior of the first font. 
We also visualize fonts that are misidentified as the same. 
In Fig.~\ref{fig:gradcam_1}~(d), the striped texture of the second font is inappropriately matching the fonts. 
From these examples, we can infer that the proposed method is able to use features such as character fill and outline to identify the fonts.

\begin{figure}[t]
	\begin{center}
    \begin{minipage}[t]{\hsize}
        \begin{center}
	        \includegraphics[width=1\textwidth]{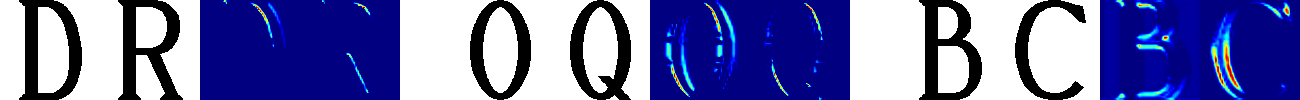}\\
	        {(a) Examples of correctly identified pairs (GT: same $\to$ prediction: same)}\\
	    \end{center}
	\end{minipage}
	\vskip 0.3cm
	\begin{minipage}[t]{\hsize}
        \begin{center}
	        \includegraphics[width=1\textwidth]{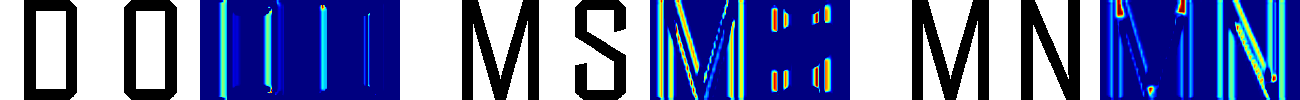}\\
	        {(b) Examples of correctly identified pairs (GT: same $\to$ prediction: same)}\\
	    \end{center}
	\end{minipage}
	\vskip 0.3cm
	\includegraphics[width=1\textwidth]{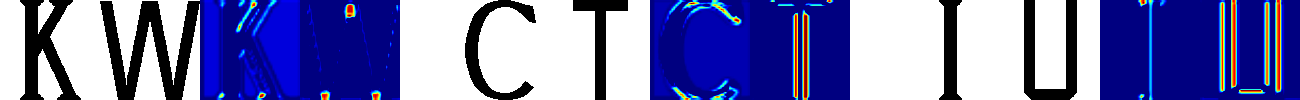}\\
		{(c) Examples of correctly identified pairs (GT: different $\to$ prediction: different)}
	\vskip 0.3cm
	\includegraphics[width=1\textwidth]{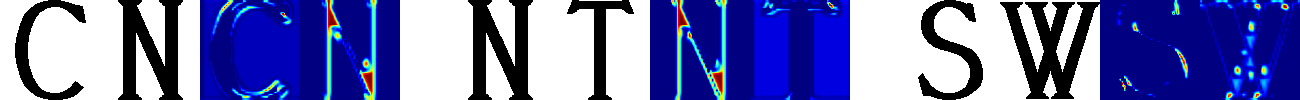}\\
		{(d) Examples of incorrectly identified pairs (GT: same $\to$ prediction: different)}\\
	\vskip 0.3cm
	\includegraphics[width=1\textwidth]{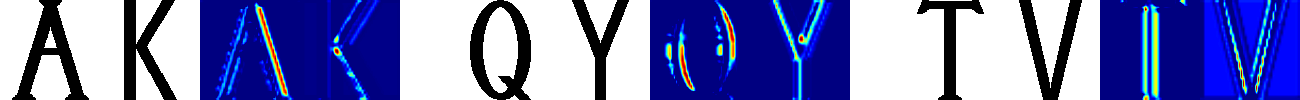}\\
		{(e) Examples of incorrectly identified pairs (GT: different $\to$ prediction: same) }
	\end{center}
    \vspace{-0.5cm} 
	\caption{Visualization of the difference between ``serif'' and ``sans serif'' font by Grad-CAM}
	\label{fig:gradcam_2}
\end{figure}

In the next example, in Fig.~\ref{fig:gradcam_2}, we demonstrate font identification between two fonts which are very similar but one having serifs and the other not having serifs.
Compared to Fig.~\ref{fig:gradcam_1}, the results of Grad-CAM in Fig.~\ref{fig:gradcam_2} show that specific regions and features are more important than overall textures. 
For example, in Fig.~\ref{fig:gradcam_2}~(a) focuses on the curves of `D,' `R,' `O,' `Q,' `B,' and `C.' 
In comparison, Fig.~\ref{fig:gradcam_2}~(b) puts importance on the vertical straight edges.

On the other hand, the font pairs correctly identified as different focus on different regions of the pairs.
Fig.~\ref{fig:gradcam_2}~(c) show the examples of contribution map which were correctly identified as different.
In this case, the top and bottom regions of the characters are highlighted. 
This is expected as the differences between the fonts should be the presence of serifs. 


As for the characters misidentified as different when the fonts were the same and same when the fonts were different, examples of Grad-CAM visualizations are shown in Figs.~\ref{fig:gradcam_2}~(d) and~(e), respectively. 
In the former case, it seems as those there were not enough common features between the two characters for the system to judge them as the same. 
For example, `C' and `S' are almost entirely composed of curves, while `N' and `W' are made of lines and angles. 
A similar phenomenon happens in Fig.~\ref{fig:gradcam_2} where the similarity of the features outweigh the differences in serif. 
In these cases, Grad-CAM shows that the serif regions are barely emphasized and the network focuses on the straight edges more.

\section{Conclusion\label{sec:conclusion}}

Character-independent font identification is a challenging task due to the differences between characters generally being greater than the differences between fonts. 
Therefore, we propose the use of a two-stream CNN-based method which determines whether two characters are from the same font or different fonts. 
As a result, we were able to demonstrate that the proposed method could identify fonts with an accuracy of 92.27$\pm$0.20\% using 6-fold cross-validation. 
This is despite using different characters as representatives of their font.

Furthermore, we perform qualitative and quantitative analysis on the results of the proposed method. 
Through the analysis, we are able to identify that the specific characters involved in the identification contributean  to the accuracy. 
This is due to certain characters containing information about the font within their native features and without common features, it is difficult for the proposed method to realize that they are the same font. 
To further support this claim, we perform an analysis on the results using PCA and Grad-CAM. 
PCA is used to show that it is easier to differentiate fonts with similar convolutional features and Grad-CAM is used to pinpoint some of the instance-wise features that led to the classifications.



In the future, we have the plan to analyze the difficulty of font identification between classes of fonts such as serif, sans serif, fancy styles, and so on.
In addition, we will try to identify fonts of other languages, including intra- and inter-language comparisons. It also might be possible to use transfer learning to identify fonts between datasets or languages. 

\appendix
\gdef\thesection{Appendix \Alph{section}}

\section{Font Identification Using a Dataset with Less Fancy Fonts}
\label{app:adobe}
The dataset used in the above experiment contains many fancy fonts and thus there was a possibility that our evaluation might overestimate the font identification performance; this is because fancy fonts are sometimes easy to be identified by their particular appearance. We, therefore, use another font dataset, called Adobe Font Folio 11.1\footnote{https://www.adobe.com/jp/products/fontfolio.html}. 
From this font set, we selected 1,132 fonts, which are comprised of 511 Serif, 314 Sans Serif, 151 Serif-Sans Hybrid, 74 Script, 61 Historical Script, and (only) 21 Fancy fonts. Note that this font type classification for the 1,132 fonts are given by \cite{Seibundo}.
We used the same neural network trained by the dataset of Section~\ref{sec:dataset}, i.e., trained with the fancy font dataset and tested on the Adobe dataset. 
Note that for the evaluation, 367,900 positive pairs and 367,900 negative pairs are prepared using the 1,132 fonts. 
Using the Adobe fonts as test, the identification accuracy was 88.33$\pm$0.89\%. 
This was lower than $92.27\%$ of the original dataset. 
However, considering the fact that formal fonts are often very similar to each other, we can still say that the character-independent font identification is possible even for the formal fonts.

\section*{Acknowledgment}
This work was supported by JSPS KAKENHI Grant Number JP17H06100.

\bibliography{hoge-nodoi} 
\bibliographystyle{splncs04}





\end{document}